\documentclass{article} 
\usepackage{iclr2027_conference,times}

\iclrfinalcopy

\usepackage[utf8]{inputenc} 
\usepackage[T1]{fontenc}    
\usepackage{hyperref}       
\usepackage{url}            
\usepackage{xurl}           
\usepackage{booktabs}       
\usepackage{amsfonts}       
\usepackage{nicefrac}       
\usepackage{microtype}      
\usepackage[table]{xcolor}  

\usepackage{amsmath,amssymb,amsthm}

\usepackage{enumitem}
\setlist{topsep=0pt, leftmargin=*}
\usepackage{appendix}

\usepackage{tabularx} 
\usepackage{booktabs} 
\usepackage{amsmath}  

\usepackage[most]{tcolorbox}
\definecolor{contribframe}{HTML}{8AAAD1}
\definecolor{contribbg}{HTML}{F5F8FC}
\newtcolorbox{mainbox}[1][]{%
  colframe=contribframe,
  colback=contribbg,
  boxrule=0.6pt,
  arc=2pt,
  breakable,
  before upper={\textbf{Main Contributions.}\par\smallskip},
  #1
}

\definecolor{fuwgreen}{RGB}{24,115,85}
\definecolor{fuwbg}{RGB}{235,247,242}
\definecolor{refbg}{RGB}{246,246,246}
\definecolor{negred}{RGB}{180,55,55}

\newif\ifincludeappendix
\ifdefined\submissiononly
  \includeappendixfalse
\else
  \includeappendixtrue
\fi
\newcommand{\appref}[1]{%
  \ifincludeappendix Appendix~\ref{#1}\else the supplementary appendix\fi}

\newtheorem{theorem}{Theorem}

\title{Token Utility for On-Policy Distillation via Local Trust-Region Optimization}

\author{
\begin{tabular}{c}
Narayanan Sadagopan$^{1}$\thanks{Corresponding author.}
\qquad
Yilan Chen$^{2}$\thanks{Work done during an internship at AWS AI Labs.}
\qquad
Vijay Lingam$^{1}$
\\[6pt]
\normalfont
$^{1}$AWS AI Labs
\qquad
$^{2}$UC San Diego
\end{tabular}
}

\begin{document}
\maketitle
\lhead{Preprint}

\begin{abstract}
Token-selection methods for on-policy distillation (OPD) prioritize student-visited positions using diverse signals, including uncertainty, teacher–student disagreement, and mismatch direction. We instead derive token importance from the distillation objective itself: how much can each position's loss decrease under an equal local KL budget on the student's predictive distribution? The answer is the utility $s_t=g_t^\top F_t^\dagger g_t$, where $g_t$ is the logit gradient and
$F_t$ is the categorical Fisher matrix. For reverse-KL it equals the student-weighted variance of the student–teacher log-probability ratio, and for forward-KL it equals the Pearson $\chi^2$ divergence, so the appropriate utility follows from the choice of objective. On shared training checkpoints, several independently motivated criteria, including student entropy, KL disagreement, and TIP, concentrate their importance on nearly the same positions as the derived utility at a 50\% selection fraction, while diverging at stricter fractions and for direction-specific criteria. We instantiate the utility as Fisher Utility Weighting (FUW), a continuous, mean-normalized token weighting. Across seven mathematical reasoning benchmarks and three Qwen3 student–teacher pairs, FUW improves over matched uniform OPD under both reverse-KL (1.35–4.0 macro points) and forward-KL (1.08–2.71 macro points), and is competitive with existing selective methods using a single closed-form score derived from the training objective.
\end{abstract}

\section{Introduction}
On-policy distillation (OPD) transfers capabilities from a stronger teacher
while retaining the student's state distribution~\citep{agarwal2023opd,gu2024minillm}. The student generates the
training trajectories, and the teacher supplies dense distributional
supervision at the prefixes that the student actually visits. This avoids the state-distribution
mismatch of offline distillation and has become an effective recipe for
improving reasoning models. Yet dense supervision does not imply that every
supervised position is equally useful. Some student and teacher distributions
are already closely aligned; elsewhere they disagree, but the correction may
require a large change in the student's predictive distribution. Uniform OPD, which weighs all token positions the same, 
does not distinguish between these cases.

Recent work addresses this heterogeneity using properties of the visited
positions. Entropy-based methods emphasize uncertain decisions
\citep{jin2026entropy}. TIP~\citep{xu2026tip} combines student uncertainty with
teacher--student divergence; DEAR~\citep{xiao2026dear} separates decision tokens
from supporting evidence tokens; Token Teachability~\citep{taopdwang2026} asks
whether teacher corrections lie within compatible student support; and TIDE~\citep{tideyu2026}
distinguishes student-excess from student-deficit probability mass. These approaches operationalize 
token importance through properties such as uncertainty, disagreement, reasoning role, support compatibility, 
or mismatch direction. We take a different approach, defining importance through the optimization problem 
itself: how much can each position's expected distillation objective improve under an equal local budget on predictive-distribution change?

Suppose that every position is allowed the same infinitesimal
change in the student's next-token distribution. Which position permits the
largest decrease in its expected distillation loss? This equal-budget comparison
separates the amount of disagreement from how effectively a local update can
resolve it. 
We formulate an independent local trust-region problem at each
student-visited prefix. Using the Fisher geometry induced by KL divergence,
the maximum first-order improvement under a common radius $\varepsilon$ is
$\sqrt{2\varepsilon s_t}$, where
\[
s_t=g_t^\top F_t^\dagger g_t.
\]
Here, $g_t$ is the gradient of the token-level distillation objective with
respect to the student logits and $F_t$ is the categorical Fisher matrix. For
reverse-KL OPD, the general score admits the closed form
\[
s_t=\operatorname{Var}_{a\sim p_{S,t}}
\left[\log\frac{p_{S,t}(a)}{p_{T,t}(a)}\right].
\]
Thus, the variance of the log-probability ratio is not introduced as a
disagreement heuristic. It is the optimization utility induced by an equal
local distributional budget constraint. We convert this score into Fisher Utility
Weighting (FUW), which applies damped, mean-normalized weights to the
token-level OPD objective. Unlike selection schemes such as TIP and TIDE, which decide whether a position contributes, 
FUW uses the utility's magnitude to set how much each position contributes, with weights proportional to 
the improvement scale $\sqrt{s_t}$. Since the utility is derived from the objective rather than chosen, 
it also provides a reference against which existing criteria can be compared. We find that several of them 
concentrate importance on largely the same positions as the derived utility, 
and that the framework accounts for where they diverge.
Our contributions are threefold:
\begin{mainbox}
\begin{itemize}
    \item We develop a common optimization framework for token utility
    in OPD. Comparing the maximum first-order loss reduction under an
    equal local KL budget yields
    $s_t=g_t^\top F_t^\dagger g_t$.
    This utility reduces exactly to student-weighted log-ratio variance
    for reverse-KL and to $\chi^2(p_T\|p_S)$ for forward-KL. (\S\ref{sec:theory}, \ref{sec:closed_form_reverse_kl}, \ref{sec:relationships}, \ref{sec:fkl})

    \item On shared checkpoints, student entropy, KL disagreement, gradient norm, TIP Soft-OR, 
    and TIDE-deficit retain nearly all of the derived utility's top-half importance mass, and conversely. 
    This suggests that these independently motivated criteria often identify similar high-importance support despite their different constructions. 
    They diverge at stricter selection fractions and for the direction-specific TIDE-excess criterion. 
    (\S\ref{sec:overlap})

    \item We turn the utility into Fisher Utility Weighting (FUW)
    and evaluate it across seven mathematical reasoning benchmarks
    and three student--teacher pairs. FUW improves macro mean accuracy
    over uniform OPD on all three pairs under the same candidate-set
    construction and loss implementation. Controlled studies examine
    the roles of score-to-position alignment, weight concentration,
    continuous weighting, and the distillation objective. (\S \ref{sec:rq1}, \ref{sec:ablation}, \ref{sec:fkl})
\end{itemize}
\end{mainbox}

\section{Theory}
\label{sec:theory}
The KL--Fisher expansion and the natural-gradient solution to a
Fisher-constrained steepest-descent problem are classical results
\citep{amari2000methods,kakade2001natural}. Our contribution is to view
token importance in on-policy distillation through the lens of local
optimization: comparing positions under a common distributional budget
yields objective-specific utilities in closed form. The trust region serves
only to derive these per-position utilities; it is not imposed as a
constraint during training.

\subsection{Problem Formulation}
On-policy distillation (OPD) trains the student on trajectories sampled from
its own policy while using a teacher model to provide token-level supervision.
For a student-generated trajectory
$x=(x_1,\ldots,x_T)\sim p_S$, define the sampled token-level reverse-KL
log-ratio at position $t$ as
\[
\ell_t
=
\log
\frac{
p_S(x_t\mid x_{<t})
}{
p_T(x_t\mid x_{<t})
}.
\]

The standard reverse-KL OPD objective is
\[
\mathcal{L}_{\mathrm{OPD}}
=
\mathbb{E}_{x\sim p_S}
\left[
\sum_{t=1}^{T} \ell_t
\right].
\]

Our goal is to determine how the optimization effort should be allocated across
the token positions in this loss. To assess the optimization opportunity at a
particular student-visited prefix $h_t=x_{<t}$, define
\[
p_{S,t}(\cdot)
=
p_S(\cdot\mid h_t),
\qquad
p_{T,t}(\cdot)
=
p_T(\cdot\mid h_t).
\]
Rather than assessing a position using only the particular token $x_t$ sampled
on the current rollout, we consider the expected token-level loss over the
student's next-token distribution. For $A_t\sim p_{S,t}$,
\[
L_t
:=
\mathbb{E}_{A_t\sim p_{S,t}}
\left[
\log
\frac{p_{S,t}(A_t)}
     {p_{T,t}(A_t)}
\right].
\]
This expectation is exactly the full-vocabulary reverse-KL:
\[
L_t
=
D_{\mathrm{KL}}
\left(
p_{S,t}\|p_{T,t}
\right).
\]

Thus, $L_t$ is the conditional expectation of the sampled token-level OPD
log-ratio,
\[
L_t = \mathbb{E}[\ell_t \mid h_t].
\]

$L_t$ marginalizes the next-token sampling randomness while keeping the student-visited prefix fixed. We use the full-distribution loss $\{L_t\}_{t=1}^{T}$ at each student visited position to determine how optimization effort
should be allocated across these positions.

Let $z_t\in\mathbb{R}^{V}$ denote the student logits at position $t$, and
define
\[
g_t
=
\nabla_{z_t}L_t.
\]
A natural measure of token importance would be the Euclidean gradient
magnitude $\|g_t\|_2$. However, this measures sensitivity per unit movement in
logit coordinates. Equal Euclidean perturbations of the logits can induce
very different changes in the corresponding predictive distributions.

We instead compare positions under an equal local change in the student's
predictive distribution. Specifically, for each position $t$, we ask for the
largest first-order reduction in the expected token-level loss that can be
achieved while restricting the local change in the student distribution:
\begin{align}
\max_{\delta_t}\quad
& -g_t^\top \delta_t
\\
\text{subject to}\quad
&
D_{\mathrm{KL}}
\left(
p_{S,t}(\cdot;z_t)
\|
p_{S,t}(\cdot;z_t+\delta_t)
\right)
\leq \varepsilon.
\end{align}

The variable $\delta_t$ is an infinitesimal logit perturbation. The constraint
compares positions by the induced change in their predictive distributions,
rather than by an arbitrary Euclidean logit displacement.

\subsection{Maximum Improvement Within the Local KL Ball}

Using the local Fisher approximation, the token-specific optimization
problem becomes
\[
\Delta_t^\star
=
\max_{\delta_t}
-g_t^\top\delta_t
\qquad
\text{subject to}
\qquad
\delta_t^\top F_t\delta_t
\le
2\varepsilon.
\]

This is the classical Fisher-constrained steepest-descent problem~\citep{kakade2001natural}. Its solution points in the natural-gradient direction:
\[
\delta_t^\star
=
-
\sqrt{
\frac{2\varepsilon}
{g_t^\top F_t^\dagger g_t}
}
F_t^\dagger g_t,
\]
where $F_t^\dagger$ is the Moore--Penrose pseudoinverse. The
pseudoinverse is required because the categorical Fisher is singular
along the constant-logit-shift direction.
Since $L_t$ depends on the logits only through softmax, $g_t$ lies in the
range of $F_t$. When $g_t=0$, we take $\delta_t^\star=0$.

The corresponding maximum first-order improvement is
\[
\Delta_t^\star
=
-g_t^\top\delta_t^\star
=
\sqrt{
2\varepsilon\,
g_t^\top F_t^\dagger g_t
}.
\]

Because every token position is evaluated under the same maximum radius $\varepsilon$, the ordering induced by $\Delta_t^\star$ is identical to the ordering induced by the normalized score
\[
s_t
:=
\frac{
(\Delta_t^\star)^2
}{
2\varepsilon
}
=
g_t^\top F_t^\dagger g_t.
\]

We call $s_t$ the \emph{local token optimization utility}. It measures the
maximum first-order improvement available at token position $t$ after
normalizing out the common maximum distributional budget. The full
Lagrangian derivation appears in \appref{app:natural-gradient}.

Importantly, this local geometry is determined entirely by the trust-region constraint on the student distribution. It does not depend on the direction of the teacher--student distillation divergence. The choice of distillation objective enters through the gradient $g$, while the local metric is determined by $F$.

\section{Closed-Form Utility for Reverse-KL OPD}
\label{sec:closed_form_reverse_kl}
We now specialize the general token utility to the reverse-KL objective used in OPD. The same framework applies to other distillation objectives through their objective-specific gradient \(g\); we derive the corresponding forward-KL utility in Appendix \ref{app:forward_kl}.

Define the vocabulary-wide log-probability ratio
\[
r(v)
=
\log\frac{p_S(v)}{p_T(v)}.
\]

\begin{theorem}[Reverse-KL Token Optimization Utility]
\label{thm:1}
For the token-level reverse-KL objective
\[
L
=
D_{\mathrm{KL}}(p_S\|p_T),
\]
the local token optimization utility satisfies
\[
s
=
g^\top F^\dagger g
=
\operatorname{Var}_{v\sim p_S}
\left[
\log\frac{p_S(v)}{p_T(v)}
\right].
\]
\end{theorem}

\paragraph{Proof sketch.}
Let $r\in\mathbb{R}^V$ denote the vector with entries
$r_v=\log(p_S(v)/p_T(v))$. Differentiating the reverse-KL with respect to the
student logits gives
\[
g = Fr,
\]
where
\[
F=\operatorname{diag}(p_S)-p_Sp_S^\top.
\]
Therefore,
\begin{align}
g^\top F^\dagger g
&=
r^\top F F^\dagger F r =r^\top F r.
\end{align}
For the categorical Fisher,
\begin{align}
r^\top F r
&=
\sum_v p_S(v)r_v^2
-
\left(\sum_vp_S(v)r_v\right)^2 \\
&=
\operatorname{Var}_{v\sim p_S}[r_v].
\end{align}
Substituting the definition of $r_v$ gives the result. A complete derivation is provided in \appref{app:reverse_kl}.

\paragraph{Practical computation.}
The utility can be computed directly in \(O(V)\) without constructing or
inverting the Fisher matrix:
\[
s=\sum_v p_S(v)\left(r_v-\sum_u p_S(u)r_u\right)^2.
\]
Since student and teacher logits are already available during distillation,
this requires only elementwise operations and reductions.

Importantly, variance is not posited as a heuristic measure of token
importance; it arises from the Fisher-constrained optimization problem.
Utility is high when student-supported actions have sufficiently different
teacher--student log-ratios to permit substantial improvement through local
probability redistribution, and low when those log-ratios are nearly constant.

\paragraph{Scope.}
The analysis is local to each student-visited predictive distribution and
does not model coupling through shared transformer parameters. Thus, the
derived utility characterizes local optimization opportunity rather than a
globally optimal weighting rule; our end-to-end experiments test whether
this signal remains useful under parameter coupling.




\section{Token Importance under Reverse-KL OPD}
\label{sec:relationships}

\paragraph{Relationship to existing criteria.}
 Existing criteria capture different notions of token importance. TIP develops
an optimization-motivated account but operationalizes its practical Soft-OR
criterion through student uncertainty and teacher--student disagreement,
while TIDE distinguishes student-excess from student-deficit mismatch. FUW
instead derives a continuous weighting signal directly from the distillation
objective through \(g_t^\top F_t^\dagger g_t\), which changes with the
objective. Table ~\ref{tab:importance_views} in Appendix~\ref{app:importance-agreement} summarizes the existing criteria
and Section \ref{sec:experiments} compares their effective support empirically.

\paragraph{Fisher Utility Weighting (FUW)}
We now turn the reverse-KL utility derived in Section \ref{sec:closed_form_reverse_kl} into a practical
training rule.
\autoref{thm:1} defines Fisher utility on the full vocabulary. For training,
we use the candidate set
\begin{equation}
\mathcal C_t
=
\operatorname{Top}_{16}(p_{S,t})
\cup
\operatorname{Top}_{16}(p_{T,t}),
\qquad |\mathcal C_t|\leq 32,
\end{equation}
and renormalize each distribution on this set. For
$m\in\{S,T\}$ and $v\in\mathcal C_t$, let
\begin{equation}
\widehat p_{m,t}(v)
=
\frac{p_{m,t}(v)}
{\sum_{u\in\mathcal C_t}p_{m,t}(u)}.
\end{equation}
We compute the restricted log-ratio, clip it, and evaluate its
student-weighted variance:
\begin{equation}
r^{\mathcal C}_t(v)
=
\log\frac{\widehat p_{S,t}(v)}{\widehat p_{T,t}(v)},
\qquad
r^{\mathrm{clip}}_t(v)
=
\operatorname{clip}\!\left(r^{\mathcal C}_t(v),-c,c\right),
\qquad
\widehat s_t
=
\operatorname{Var}_{v\sim\widehat p_{S,t}}
\!\left[r^{\mathrm{clip}}_t(v)\right].
\label{eq:practical_score}
\end{equation}
We set $c=5$, so $0\leq\widehat s_t\leq c^2$.
Without clipping, $\widehat s_t$ is the Fisher utility of the
reverse-KL loss on $\mathcal C_t$, with the candidate set held fixed.
Clipping stabilizes this score and makes it an approximation to that
utility.

The per-position training loss is the reverse-KL on the same
renormalized candidate set:
\begin{equation}
\widehat L_t
=
D_{\mathrm{KL}}\!\left(
\widehat p_{S,t}\middle\|\widehat p_{T,t}
\right).
\label{eq:restricted_loss}
\end{equation}
For the $N$ valid positions in a batch, we set
$u_t=\sqrt{\widehat s_t}$ and use mean-one weights
\begin{equation}
w_t=
\begin{cases}
\displaystyle \frac{N u_t}{\sum_{j=1}^{N}u_j},
    & \sum_{j=1}^{N}u_j>0,\\[7pt]
1, & \sum_{j=1}^{N}u_j=0.
\end{cases}
\label{eq:fuw_weights}
\end{equation}
The resulting objective is
\begin{equation}
\mathcal L_{\mathrm{FUW}}
=
\frac{1}{N}\sum_{t=1}^{N}
\operatorname{stopgrad}(w_t)\widehat L_t .
\label{eq:fuw_loss}
\end{equation}
Uniform OPD uses the same candidate set, renormalization, and
per-position loss, with $w_t=1$ for every position. Thus, its
comparison with FUW changes the weighting rule while holding this
loss implementation fixed.

The square-root transform follows from the equal-radius improvement scale,
\(\Delta_t^\star\propto\sqrt{s_t}\), and also dampens score concentration.
We therefore use \(u_t=\sqrt{\hat{s}_t}\) as a theoretically motivated
weighting signal. We normalize these utilities to have unit mean and treat
the resulting weights as stop-gradient during optimization. Unless otherwise
specified, Fisher Utility Weighting (FUW) refers to this reverse-KL
instantiation; Section~\ref{sec:fkl} applies the same construction to
forward KL using its corresponding objective-specific utility.

\section{Experiments}
\label{sec:experiments}

Our experiments evaluate the derived token utility from these complementary
perspectives: end-to-end distillation performance, its relationship to existing
token-importance criteria, and controlled variations of the scoring and weighting
procedure. We evaluate both reverse- and forward-KL OPD across three
student--teacher pairs and seven mathematical-reasoning benchmarks, using the
objective-specific utility induced by each distillation objective. Additional
concentration, robustness, and diagnostic analyses are reported in \appref{app:additional_studies}.

\subsection{RQ1: Does FUW improve over uniform reverse-KL OPD across model pairs?}
\label{sec:rq1}
\begin{figure*}[t]
  \centering
  \includegraphics[width=\textwidth]{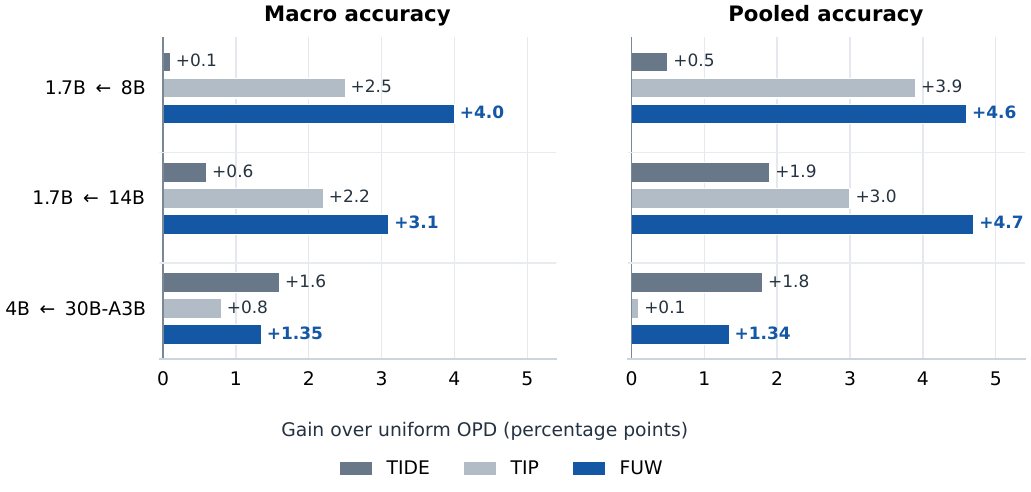}
  \caption{Macro and pooled average gains over uniform OPD across three
  Qwen3 student--teacher pairs. Bars show percentage-point differences
  in the reported point estimates. Macro weights the seven benchmarks
  equally; pooled weights all 1,620 problems equally. Uniform OPD and
  FUW use the same candidate-set construction and restricted reverse-KL
  loss, differing in their token weights. All methods share the same training setup. 
  Uniform OPD, TIP, and FUW use the same candidate-set construction; TIDE uses its native support (Appendix~\ref{app:main_table_exps}).}
  \label{fig:cross_pair_results}
\end{figure*}

We evaluate reverse-KL OPD on three Qwen3 student--teacher pairs:
Qwen3-1.7B-Base $\leftarrow$ Qwen3-8B,
Qwen3-1.7B-Base $\leftarrow$ Qwen3-14B, and
Qwen3-4B-Base $\leftarrow$ Qwen3-30B-A3B-Instruct-2507.
For all model pairs, training uses DAPO-Math-17k~\citep{dapo} for
279 steps (one epoch), 256 sequences per step, and an 8,192 tokens maximum training context with maximum response length set to 7,168 tokens and maximum prompt length set to 1,024 tokens.
FUW uses $w_t\propto\sqrt{\widehat{s}_t}$.
Figure~\ref{fig:cross_pair_results} summarizes aggregate performance;
per-benchmark results along with details of the evaluation setup and baselines are in Appendix~\ref{app:main_table_exps}.

\paragraph{Benchmarks.}

We evaluate on 1,620 problems from seven mathematical-reasoning
benchmarks: AIME24 (30), AIME25 (30), AMC23 (83), BRUMO25 (30),
MATH-500 (500), Minerva (272), and OlympiadBench (675).
AIME24/25, AMC23, and BRUMO25 form the four smaller benchmarks. 
Evaluations use Macro Avg@8, which weights benchmarks equally, and pooled Avg@8, which weights problems equally. 

\paragraph{Results.}
FUW improves on uniform OPD on all three pairs. Its Macro Avg@8 gains
are $4.0$, $3.1$, and $1.35$ percentage points for the
1.7B $\leftarrow$ 8B, 1.7B $\leftarrow$ 14B, and
4B $\leftarrow$ 30B-A3B pairs, respectively. The corresponding
pooled gains are $4.6$, $4.7$, and $1.34$ points. Thus, the gain over
uniform OPD persists across the evaluated model pairs, although its
size varies.

On the 1.7B $\leftarrow$ 8B pair, FUW achieves the highest aggregate
scores among the trained students: $26.8$ Macro Avg@8 and $39.2$
pooled Avg@8. It exceeds TIP Soft-OR by $1.5$ and $0.7$ points,
respectively, and has the highest point estimate on five of the seven
individual benchmarks among the trained methods. The pooled
FUW--TIP difference is small: its paired-bootstrap 95\% confidence
interval includes zero. 


FUW also has the highest aggregate point estimates among trained
methods on the 1.7B $\leftarrow$ 14B pair. On the
4B $\leftarrow$ 30B-A3B pair, TIDE is slightly ahead: its Macro and
pooled Avg@8 are $46.4$ and $58.0$, compared with FUW's $46.15$
and $57.54$. Thus, FUW consistently improves over matched-implementation
uniform OPD, while its differences from selective baselines are smaller
and less uniform.

\subsection{RQ2: Do different token-importance criteria prioritize the
same positions?}
\label{sec:overlap}
We compare FUW with Euclidean gradient magnitude, reverse-KL
disagreement, student entropy, TIP Soft-OR, and the excess and deficit
components of TIDE; their definitions and relationships to FUW are
summarized in Table \ref{tab:importance_views} in Appendix \ref{app:importance-agreement}. We compute all these token importance criteria on the checkpoints 
obtained during uniformly weighted reverse-KL OPD of Qwen3-1.7B-Base $\leftarrow$ Qwen3-14B. Since we use uniformly weighted OPD the token-importance criteria do not determine the model states being compared.

Different importance criteria may disagree on the exact ordering of positions while still concentrating their weight on similar positions of the trajectory. We therefore measure agreement in both directions. 
Let \(S_X^\rho\) denote the top-\(\rho\) fraction of positions selected by criterion \(X\), and let \(w_{X,i}\) denote its weight on position \(i\). We define
$$ \mathrm{REV}_\rho(X) = \frac{\sum_{i\in S_X^\rho}w_{\mathrm{FUW},i}} {\sum_{i\in S_{\mathrm{FUW}}^\rho}w_{\mathrm{FUW},i}}, \qquad \mathrm{CONT}_\rho(X) = \frac{\sum_{i\in S_{\mathrm{FUW}}^\rho}w_{X,i}} {\sum_{i\in S_X^\rho}w_{X,i}}. $$

REV asks how much of FUW's attainable top-\(\rho\) weight is retained when positions are selected by \(X\), while CONT asks the converse. We summarize the two directions using
$$ B_\rho(X) = \min\{\mathrm{REV}_\rho(X),\mathrm{CONT}_\rho(X)\}, $$
which we call bidirectional weight retention. Importantly, high \(B_\rho\) measures agreement in effective high-importance support, not equality of the underlying weight vectors. Two criteria can attain \(B_\rho\approx1\) while assigning different relative weights to individual positions within that shared support. 


\begin{figure*}[t]
    \centering
    \includegraphics[width=\textwidth]{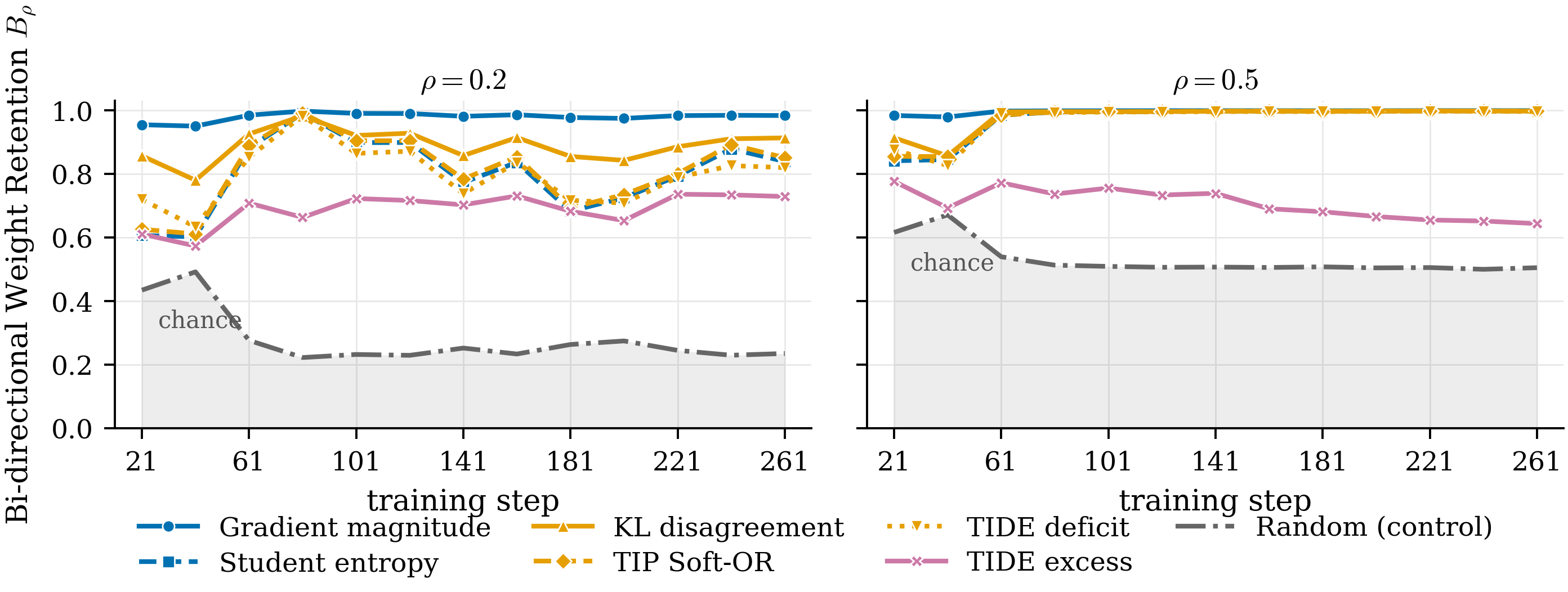}
    \caption{Effective-support agreement \(B_\rho\) with FUW during uniform reverse-KL OPD of Qwen3-1.7B-Base $\leftarrow$ Qwen3-14B. At \(\rho=0.2\), agreement is criterion- and checkpoint-dependent; at \(\rho=0.5\), most criteria approach near-complete retention, while TIDE-excess remains distinct. The random-selection control accounts for agreement induced by the concentration of FUW's weights.}
    \label{fig:wt_equiv}
\end{figure*}


We evaluate \(\rho=0.2\), close to TIDE's realized active fraction (0.264), and \(\rho=0.5\), matching TIP's 50\% Soft-OR gate. As shown in Figure \ref{fig:wt_equiv}, at \(\rho=0.2\), agreement is substantial but criterion- and checkpoint-dependent, with gradient magnitude remaining particularly close to FUW. At \(\rho=0.5\), gradient magnitude, student entropy, KL disagreement, TIP Soft-OR, and TIDE-deficit achieve \(B_{0.5}\geq0.985\) from step 61 onward, whereas TIDE-excess remains distinct. Thus, for the Qwen3-1.7B-Base \(\leftarrow\) Qwen3-14B pair, several
independently motivated criteria identify nearly the same effective
high-importance support at the broader operating fraction, although they may
assign different relative weights within that support. Appendix
\ref{app:importance-agreement} further analyzes \(B_\rho(X)\) into its directional components $\mathrm{REV}_\rho(X)$ and $\mathrm{CONT}_\rho(X)$ 
and defines the random-control, showing that the near-unit \(B_{0.5}\) values for
most criteria reflect nearly symmetric agreement, whereas TIDE-excess exhibits
asymmetric containment. It also reports a cross-pair robustness analysis on
Qwen3-1.7B-Base \(\leftarrow\) Qwen3-8B along an FUW-trained trajectory; although
these checkpoints are not independent of the reference criterion, the same
qualitative effective-support structure appears at both \(\rho=0.2\) and
\(0.5\).

\subsection{RQ3: Which components of FUW matter?}
\label{sec:ablation}
These ablations use reverse-KL distillation on the Qwen3-1.7B-Base \(\leftarrow\) Qwen3-8B pair with the same training setup as RQ1. We separate the roles of the metric, position-specific utility information, and weighting form. Replacing \(g_t^\top F_t^\dagger g_t\) with the Euclidean score \(\|g_t\|_2^2\) gives comparable downstream accuracy, although its weights become substantially more concentrated during training: effective retention falls to \(0.104\) and the maximum token weight reaches \(96\). FUW avoids this late-stage concentration, although the two metrics are not separated by downstream accuracy at the present training budget. Permuting FUW weights within each batch, while preserving their marginal distribution, lowers pooled Avg@8 by \(1.02\) points (\(p=0.028\); Table \ref{tab:weighting_controls}), showing that position-specific utility information matters. In contrast, a hard top-half FUW mask reaches \(38.4\) pooled Avg@8, compared with \(39.2\) for continuous FUW and \(38.4\) for TIP Soft-OR at the same 50\% retention fraction. This is consistent with RQ2, where FUW and TIP identify nearly the same effective support at \(\rho=0.5\). The complete paired contrasts appear in \appref{app:weighting_controls}.


\subsection{RQ4: How does the token-utility framework extend to forward-KL OPD?}
\label{sec:fkl}

The token-utility framework extends directly to forward-KL distillation:
applying the same local trust-region formulation yields the Pearson
$\chi^2(p_T\|p_S)$ divergence as the induced utility, rather than the log-ratio
variance obtained for reverse-KL. A complete derivation of the forward-KL utility is provided in \appref{app:forward_kl}.
For practical weighting, we upper-clip the log-ratio
\(\tilde r=\log(p_T/p_S)\) at \(\tau=3\), i.e.,
\(p_T/p_S\le e^\tau\approx20\), when computing the detached utility
weights. The utility is computed on the same candidate set \(C_t\), and
weights are formed as in Eqs. (10)–(11). The forward-KL loss itself is
unchanged.

We test whether weighting forward-KL OPD using its corresponding utility improves over uniform forward-KL distillation. 
We use the same experimental setup as in RQ1, changing only the objective-specific components: the forward-KL loss and its corresponding Pearson \(\chi^2(p_T\|p_S)\) utility. Uniform forward-KL uses the same forward-KL loss with uniform token weights. Objective-specific utility weighting improves both Macro and pooled Avg@8 on all three model pairs. Averaged equally across the three model pairs, utility weighting improves Macro Avg@8 by 1.65 points and pooled Avg@8 by 1.95 points. Detailed results are shown in Table ~\ref{tab:fkl-pass8} in Appendix \ref{app:fkl_results}.


Thus, the same local trust-region principle developed in Section \ref{sec:theory} yields a usable utility for a different objective: the Pearson \(\chi^2(p_T\|p_S)\) divergence for forward-KL, in place of the log-ratio variance for reverse-KL. Weighting by the corresponding objective-specific utility improves aggregate Avg@8 over uniform weighting under both reverse- and forward-KL distillation.

\section{Related Work}

\paragraph{On-policy distillation.}
On-policy distillation (OPD) trains a student model on trajectories sampled from its own policy while receiving token-level supervision from
a stronger teacher~\citep{agarwal2023opd,gu2024minillm}. Recent work has
demonstrated that OPD substantially improves reasoning performance over offline distillation, while simultaneously exposing new optimization
challenges unique to student-generated trajectories. \citet{luo2026demystifying} attribute instability to repetitive-token
advantage amplification, while \citet{li2026rethinking}
characterize failure modes arising from signal degradation on drifted
prefixes. Other work has proposed stabilizing modifications including
reward extrapolation~\citep{yang2026generalized}, relaxed clipping strategies~\citep{ko2026scaling}, and entropy-aware training recipes
\citep{jin2026entropy}. Collectively, these studies suggest that the
distillation signal is highly non-uniform across token positions and
motivate selective supervision strategies.

\paragraph{Token importance and selective supervision.}
Entropy-based methods prioritize uncertain positions
\citep{wang2026beyond80,jin2026entropy}. TIP~\citep{xu2026tip} combines uncertainty and
teacher--student divergence; DEAR~\citep{xiao2026dear} distinguishes decision and
supporting evidence tokens; Token Teachability~\citep{taopdwang2026} measures
whether disagreement lies within compatible student support; and 
TIDE~\citep{tideyu2026} separates student-excess from student-deficit
mass. REOPOLD~\citep{ko2026scaling} instead treats token log-ratios as rewards and
combines clipping with entropy-guided sampling. FUW is
complementary: it derives a score from achievable loss reduction under a
common distributional budget rather than prescribing a token property.

\paragraph{Information geometry and optimization-based token utility.}
Our work is based on the information-geometric view of optimization. The
local Fisher geometry induced by KL divergence and the resulting
natural-gradient formulation have a long history in machine learning
\citep{amari2000methods,kakade2001natural}. Trust-region methods such as
TRPO similarly constrain policy updates using KL divergence
\citep{schulman2015trpo}, while PPO provides a practical approximation to
trust-region policy optimization~\citep{schulman2017ppo}. Rather than
applying these ideas to a shared parameter-space update, we use the
local KL geometry as an analytical lens for characterizing individual
student-visited positions. Specifically, we define token utility as the maximum first-order reduction in 
the token-level distillation objective under a common local budget on predictive-distribution change.
This yields the Fisher-normalized quantity $g_t^\top F_t^\dagger g_t$, which reduces to student-weighted log-ratio
variance for reverse KL and Pearson $\chi^2(p_T\|p_S)$ divergence for
forward KL.
This provides a complementary perspective to recent notions based on
uncertainty, disagreement, reasoning role, support compatibility, or
mismatch type: rather than specifying a property that makes a token informative
or learnable, we define the local optimization problem and let utility emerge
from its solution.

\section{Discussion and Conclusion}
We derived token utility for on-policy distillation (OPD) from the 
objective being optimized. The local trust-region formulation defines
utility through the optimization opportunity induced by the
distillation objective, yielding the general quantity
$g_t^\top F_t^\dagger g_t$. Under reverse-KL this reduces to the
student-weighted variance of the log-probability ratio, while under
forward-KL it becomes the Pearson $\chi^2(p_T\|p_S)$ divergence.
Across three student--teacher pairs, weighting by the corresponding
utility improves aggregate performance over uniform weighting under
both objectives.

Distinct importance criteria can nevertheless identify similar
effective high-importance support. Under reverse-KL OPD, several
criteria with different motivations exhibit near-complete
bidirectional retention with FUW at the broader operating fraction,
while agreement is more criterion- and checkpoint-dependent at the
more selective fraction and TIDE-excess remains distinct. Importantly,
shared effective support does not imply identical weighting behavior,
as the criteria can distribute importance differently within that
support.

Our analysis is local and first-order: it evaluates predictive
distributions independently at fixed student-visited prefixes, whereas
shared transformer parameters couple positions during training.
Accordingly, the derived utility characterizes local optimization
opportunity rather than a globally optimal shared-parameter weighting
rule. Our experiments are restricted to Qwen3 model pairs and mathematical reasoning, leaving
broader architectures and agentic tasks for future study.

Overall, the framework provides an optimization-based construction for
token utility in OPD: specify the distillation objective, derive its
local utility under a common predictive-distribution budget, and use
that utility to characterize or prioritize student-visited positions.

\section{Acknowledgments}
The authors thank Prof. Tong Zhang of UIUC for his valuable feedback on an earlier version of this work.

\clearpage

\section*{AI Use Statement}
The authors used LLM assistants to edit prose, check consistency between
tables and text, and implement plotting functions for figures. The authors
designed the experiments, verified every reported number against the run
artifacts, and take full responsibility for the content.

\section*{Ethics Statement}
This work studies token weighting for on-policy distillation of language
models on publicly available mathematical-reasoning data and benchmarks.
It involves no human subjects or personal data. Distillation transfers
capabilities from a larger teacher to a smaller student; the student may
therefore inherit the teacher's errors and biases, and more efficient
distillation lowers the cost of deploying capable models. We evaluate only
mathematical reasoning and make no claims about safety-relevant behavior
of the distilled students.

\section*{Reproducibility Statement}
Section~\ref{sec:relationships} specifies the Fisher Utility Weighting (FUW) score, candidate-set
construction, clipping, and weighting. Section~\ref{sec:rq1} and
Appendix \ref{app:main_table_exps} give the model pairs, training data, training budget, sampling settings, and evaluation protocol. The Appendix ~\ref{app:theory} contains complete derivations for all theoretical results. Each configuration is trained once; reported uncertainty reflects paired bootstrap resampling over evaluation problems. For every run we retain the configuration, intermediate checkpoints, and raw evaluation outputs. We will release code
and configurations upon publication.

\bibliographystyle{iclr2027_conference}
\bibliography{references}

\ifincludeappendix
\newpage
\appendix
\onecolumn
\appendixpage

\section{Additional Theoretical Derivations}
\label{app:theory}

\subsection{Derivation of the Fisher-Metric Constraint}
\label{app:fisher_constraint}

In the main text we constrain the amount by which the student's predictive distribution changes,

\begin{equation*}
D_{\mathrm{KL}}
\!\left(
p_S(\cdot;z)
\,\middle\|\,
p_S(\cdot;z+\delta)
\right)
\le
\varepsilon.
\end{equation*}

This appendix derives the corresponding local quadratic approximation.

Define

\begin{equation*}
C(\delta)
=
D_{\mathrm{KL}}
\!\left(
p_S(\cdot;z)
\,\middle\|\,
p_S(\cdot;z+\delta)
\right).
\end{equation*}

Using the definition of KL divergence,

\begin{equation*}
C(\delta)
=
\sum_i
p_S(i;z)
\log
\frac
{p_S(i;z)}
{p_S(i;z+\delta)}.
\end{equation*}

Since the first term is independent of $\delta$,

\begin{equation*}
C(\delta)
=
-
\sum_i
p_S(i;z)
\log p_S(i;z+\delta)
+
\mathrm{const}.
\end{equation*}

Taking the derivative with respect to $\delta_j$,

\begin{equation*}
\frac{\partial C}{\partial\delta_j}
=
-
\sum_i
p_S(i)
\frac{\partial}
{\partial\delta_j}
\log p_S(i;z+\delta).
\end{equation*}

Using the softmax identity

\[
\frac{\partial\log p(i)}
{\partial z_j}
=
\mathbf{1}[i=j]-p(j),
\]

gives

\begin{equation*}
\frac{\partial C}{\partial\delta_j}
=
p_S(j;z+\delta)
-
p_S(j;z).
\end{equation*}

Evaluating at the expansion point,

\[
\delta=0,
\]

yields

\begin{equation*}
\nabla C(0)=0.
\end{equation*}

Similarly,

\[
C(0)=0.
\]

Therefore,

\[
C(\delta)
=
\frac12
\delta^\top
H
\delta
+
O(\|\delta\|^3),
\]

where

\[
H
=
\nabla_\delta^2
C(0).
\]

Differentiating once more,

\[
H_{ij}
=
\frac{\partial p_S(j)}
{\partial z_i}.
\]

Using the softmax derivative,

\[
\frac{\partial p(j)}
{\partial z_i}
=
p(j)
(\mathbf{1}[i=j]-p(i)),
\]

gives

\[
H
=
\operatorname{diag}(p_S)
-
p_Sp_S^\top.
\]

Recognizing this as the Fisher information matrix~\citep{amari2000methods},

\[
F
=
\operatorname{diag}(p_S)
-
p_Sp_S^\top,
\]

we obtain

\begin{equation*}
D_{\mathrm{KL}}
\!\left(
p_S(\cdot;z)
\,\middle\|\,
p_S(\cdot;z+\delta)
\right)
=
\frac12
\delta^\top
F
\delta
+
O(\|\delta\|^3),
\end{equation*}

which is the local Fisher-metric constraint used in the main text.

\subsection{Solution of the Fisher-Constrained Optimization}
\label{app:natural-gradient}

We derive the solution to
\begin{equation*}
\max_{\delta}\; -g^\top\delta
\qquad
\text{subject to}
\qquad
\delta^\top F\delta\le 2\varepsilon.
\end{equation*}

Here $F$ is the Fisher information matrix of the categorical student
distribution. The derivation follows the classical Fisher-constrained steepest-descent argument of natural gradient methods~\citep{kakade2001natural}. Since the softmax is invariant to constant logit shifts,
$F$ is positive semi-definite rather than positive definite. Its null
space contains the all-ones vector $\mathbf{1}$. We therefore restrict
the optimization to the simplex tangent space
\begin{equation*}
\mathcal{T}
=
\left\{
\delta\in\mathbb{R}^{V}:
\mathbf{1}^\top\delta=0
\right\},
\end{equation*}
on which the Fisher is positive definite when the student distribution
has full support. Equivalently, its inverse action in the ambient logit
space is represented by the Moore--Penrose pseudoinverse $F^\dagger$.

Because the objective is linear and the feasible set is convex, a
nonzero optimum lies on the boundary:
\begin{equation*}
\delta^\top F\delta=2\varepsilon.
\end{equation*}

Introduce a Lagrange multiplier $\lambda$ and define
\begin{equation*}
\mathcal{J}(\delta,\lambda)
=
-g^\top\delta
-
\lambda
\left(
\delta^\top F\delta-2\varepsilon
\right).
\end{equation*}

The sign convention for $\lambda$ is arbitrary; we choose it so that
the resulting optimizer points in the descent direction. The
first-order stationarity condition with respect to $\delta$ is
\begin{align*}
\nabla_\delta\mathcal{J}
&=
-g-2\lambda F\delta
=
0.
\end{align*}

Hence,
\begin{equation*}
F\delta
=
-\frac{1}{2\lambda}g.
\end{equation*}

Since the gradient of the categorical KL satisfies
$\mathbf{1}^\top g=0$, it lies in the range of $F$. Applying the
pseudoinverse gives
\begin{equation*}
\delta
=
-\frac{1}{2\lambda}F^\dagger g
+
u,
\end{equation*}
where $u$ lies in the null space of $F$. A null-space component changes
all logits by a common shift and therefore changes neither the softmax
distribution nor the objective. We select the minimum-norm
representative by taking $u=0$:
\begin{equation}
\delta
=
-\frac{1}{2\lambda}F^\dagger g.
\label{eq:scaled-natural-gradient}
\end{equation}

Substituting Equation~\eqref{eq:scaled-natural-gradient} into the active
constraint gives
\begin{align*}
2\varepsilon
&=
\delta^\top F\delta
\\
&=
\frac{1}{4\lambda^2}
g^\top
F^\dagger F F^\dagger
g.
\end{align*}

Because $g$ lies in the range of $F$,
\[
F F^\dagger g=g,
\]
and therefore
\begin{equation*}
2\varepsilon
=
\frac{1}{4\lambda^2}
g^\top F^\dagger g.
\end{equation*}

Solving for the positive multiplier magnitude,
\begin{equation*}
\lambda
=
\frac{1}{2}
\sqrt{
\frac{g^\top F^\dagger g}
{2\varepsilon}
}.
\end{equation*}

Substituting back into
Equation~\eqref{eq:scaled-natural-gradient} yields
\begin{equation*}
\boxed{
\delta^\star
=
-
\sqrt{
\frac{2\varepsilon}
{g^\top F^\dagger g}
}
F^\dagger g
}.
\end{equation*}

Thus, the optimal perturbation points in the natural-gradient direction
$-F^\dagger g$, with its scale chosen so that it exhausts the permitted
local distributional budget.

Finally, its first-order improvement is
\begin{align*}
-g^\top\delta^\star
&=
\sqrt{
\frac{2\varepsilon}
{g^\top F^\dagger g}
}
g^\top F^\dagger g
\\
&=
\boxed{
\sqrt{
2\varepsilon\,
g^\top F^\dagger g
}
}.
\end{align*}

For a common budget $\varepsilon$ across token positions, the factor
$2\varepsilon$ and the square root are common monotone transformations.
Consequently, ranking tokens by maximum local first-order improvement
is equivalent to ranking them by
\begin{equation*}
g^\top F^\dagger g.
\end{equation*}

\subsection{Closed Form for Reverse-KL Token Utility}
\label{app:reverse_kl}

For a fixed student-visited prefix, let
\[
p_i=p_S(i), \qquad q_i=p_T(i),
\]
and define
\[
r_i=\log\frac{p_i}{q_i}.
\]
The reverse-KL objective is
\[
L
=
D_{\mathrm{KL}}(p\|q)
=
\sum_i p_i\log\frac{p_i}{q_i}
=
\sum_i p_i r_i.
\]

We first derive its gradient with respect to the student logits $z$.
For the softmax distribution,
\[
\frac{\partial p_i}{\partial z_j}
=
p_i(\mathbf{1}[i=j]-p_j).
\]
Moreover,
\[
\frac{\partial}{\partial p_i}
\left(
p_i\log\frac{p_i}{q_i}
\right)
=
\log\frac{p_i}{q_i}+1
=
r_i+1.
\]
Applying the chain rule,
\begin{align*}
g_j
=
\frac{\partial L}{\partial z_j}
&=
\sum_i
(r_i+1)
\frac{\partial p_i}{\partial z_j}
\\
&=
\sum_i
(r_i+1)
p_i(\mathbf{1}[i=j]-p_j)
\\
&=
p_j(r_j+1)
-
p_j\sum_i p_i(r_i+1).
\end{align*}
Since $\sum_i p_i=1$,
\[
\sum_i p_i(r_i+1)
=
\sum_i p_i r_i+1.
\]
Therefore,
\begin{align*}
g_j
&=
p_j
\left(
r_j-\sum_i p_i r_i
\right)
\\
&=
p_j
\left(
r_j-\mathbb{E}_{i\sim p}[r_i]
\right).
\end{align*}

Now recall that the categorical Fisher information matrix is
\[
F
=
\operatorname{diag}(p)-pp^\top.
\]
Multiplying $F$ by $r$ gives
\[
(Fr)_j
=
p_jr_j-p_j\sum_i p_i r_i,
\]
which is exactly the expression above for $g_j$. Hence,
\[
\boxed{g=Fr.}
\]

The Fisher-constrained token utility derived in the main text is
\[
s=g^\top F^\dagger g.
\]
Substituting $g=Fr$,
\begin{align*}
s
&=
(Fr)^\top F^\dagger(Fr)
\\
&=
r^\top FF^\dagger Fr
\\
&=
r^\top Fr,
\end{align*}
where we use the Moore--Penrose identity
\[
FF^\dagger F=F.
\]

Expanding the categorical Fisher,
\begin{align*}
r^\top Fr
&=
r^\top
\left(
\operatorname{diag}(p)-pp^\top
\right)r
\\
&=
\sum_i p_i r_i^2
-
\left(\sum_i p_i r_i\right)^2
\\
&=
\mathbb{E}_{i\sim p}[r_i^2]
-
\mathbb{E}_{i\sim p}[r_i]^2
\\
&=
\operatorname{Var}_{i\sim p}[r_i].
\end{align*}
Substituting
$r_i=\log(p_S(i)/p_T(i))$ therefore yields
\[
\boxed{
s
=
\operatorname{Var}_{i\sim p_S}
\left[
\log\frac{p_S(i)}{p_T(i)}
\right].
}
\]

\subsection{Closed Form for Forward-KL Token Utility}
\label{app:forward_kl}

The local Fisher geometry derived in Appendix~A.1 depends only on the
constraint on changes to the student distribution. Changing the
distillation objective therefore changes the utility through its gradient
$g$, while the Fisher metric remains the same.

Let $p=p_S$ and $q=p_T$. For the forward-KL objective,
\[
L_{\mathrm{FKL}}
=
D_{\mathrm{KL}}(q\|p)
=
\sum_i q_i \log \frac{q_i}{p_i}.
\]
Differentiating with respect to the student logits gives
\[
g = p-q.
\]
Since $p$ and $q$ are probability distributions,
\[
\mathbf{1}^\top g = 0.
\]

From Appendix~A.2, the local token utility is
\[
s_{\mathrm{FKL}} = g^\top F^\dagger g,
\qquad
F=\operatorname{diag}(p)-pp^\top.
\]
Let $D=\operatorname{diag}(p)$. Since
$p^\top D^{-1}=\mathbf{1}^\top$,
\[
FD^{-1}g
=
(D-pp^\top)D^{-1}g
=
g-p(\mathbf{1}^\top g)
=
g.
\]
Thus $D^{-1}g$ is a solution to the Fisher system $Fv=g$.
Any two solutions differ only by a null-space component of $F$.
As established in Appendix~A.2, this component is a constant-logit
shift and is orthogonal to $g$. Therefore,
\[
g^\top F^\dagger g
=
g^\top D^{-1}g
=
\sum_i \frac{g_i^2}{p_i}.
\]
Substituting $g_i=p_i-q_i$ gives
\[
\boxed{
s_{\mathrm{FKL}}
=
\sum_i \frac{(q_i-p_i)^2}{p_i}
=
\chi^2(q\|p)
}.
\]

Thus, under the same local predictive-distribution geometry, reverse-KL
and forward-KL induce different token utilities: reverse-KL yields the
student-weighted variance of the log-probability ratio, whereas forward
KL yields the Pearson $\chi^2$ divergence.

\subsection{Exact Finite-Radius Utility for Reverse-KL}
\label{app:exact_finite_radius}

Relation to the local Fisher utility. The preceding derivations define token utility through a local quadratic approximation to the predictive-distribution KL constraint. We now ask whether this quantity also corresponds to an exact distribution-space optimization problem before taking the local approximation. For reverse-KL, we solve the finite-radius problem exactly and show that the Fisher utility is recovered as its normalized infinitesimal limit.

At one student-visited prefix, write
\[
p_i=p_S(i;z),
\qquad
q_i=p_T(i),
\qquad
x_i(\delta)=p_S(i;z+\delta),
\]
and assume $p_i,q_i>0$. Since softmax can represent every distribution in the
interior of the simplex, optimizing over $\delta$ up to a common logit shift is
equivalent to
\begin{equation*}
\min_{x}
\quad
D_{\mathrm{KL}}(x\|q)
\qquad
\text{subject to}
\qquad
D_{\mathrm{KL}}(p\|x)\leq\varepsilon,
\quad
\sum_i x_i=1.
\end{equation*}
The objective and feasible set are convex in $x$, so the optimizer is unique.
If $D_{\mathrm{KL}}(p\|q)\leq\varepsilon$, then $x^\star=q$. Otherwise, the
constraint is active.

For the active case, introduce multipliers $\lambda>0$ and $\nu$:
\begin{align*}
\mathcal{J}(x,\lambda,\nu)
&=
\sum_i x_i\log\frac{x_i}{q_i}
+
\lambda
\left(
\sum_i p_i\log\frac{p_i}{x_i}
-\varepsilon
\right)
+
\nu\left(\sum_i x_i-1\right).
\end{align*}
Stationarity with respect to $x_i$ gives
\begin{equation}
\log\frac{x_i}{q_i}
+1+\nu
-
\lambda\frac{p_i}{x_i}
=0.
\label{eq:exact_finite_stationarity}
\end{equation}
Let $u_i=\lambda p_i/x_i$ and
$\alpha=\lambda e^{1+\nu}$. Equation~\eqref{eq:exact_finite_stationarity}
then implies
\[
u_i e^{u_i}
=
\alpha\frac{p_i}{q_i}.
\]
Using the principal Lambert function $W_0$, defined by
$W_0(y)e^{W_0(y)}=y$, yields
\[
u_i
=
W_0\!\left(\alpha\frac{p_i}{q_i}\right).
\]
Normalizing $x$ therefore gives
\begin{equation}
\boxed{
x_i^\star(\alpha)
=
\frac{
p_i/
W_0\!\left(\alpha p_i/q_i\right)
}{
\displaystyle
\sum_j
p_j/
W_0\!\left(\alpha p_j/q_j\right)
}
}.
\label{eq:exact_finite_distribution}
\end{equation}
The scalar $\alpha>0$ is chosen so that
\begin{equation}
D_{\mathrm{KL}}
\left(
p\middle\|x^\star(\alpha)
\right)
=
\varepsilon.
\label{eq:exact_finite_radius}
\end{equation}
As $\alpha\downarrow0$, $x^\star(\alpha)\to q$, while as
$\alpha\uparrow\infty$, $x^\star(\alpha)\to p$. Thus
Equation~\eqref{eq:exact_finite_radius} is a one-dimensional root problem for
$0<\varepsilon<D_{\mathrm{KL}}(p\|q)$.

Because
$x_i\propto p_i e^{\delta_i}$, an exact optimal logit perturbation is
\begin{equation*}
\boxed{
\delta_i^\star
=
-
\log
W_0\!\left(\alpha\frac{p_i}{q_i}\right)
+c
},
\end{equation*}
where the common shift $c$ does not affect the softmax distribution.

The exact finite-radius improvement is
\begin{equation*}
U(\varepsilon)
:=
D_{\mathrm{KL}}(p\|q)
-
D_{\mathrm{KL}}(x^\star\|q).
\end{equation*}
To connect it to the local utility, let $r_i=\log(p_i/q_i)$. As
$\varepsilon\downarrow0$, the exact optimizer satisfies
$x^\star=p+O(\sqrt{\varepsilon})$. The objective and constraint expansions in
logit space give
\begin{align*}
D_{\mathrm{KL}}(p\|x(\delta))
&=
\frac12\delta^\top F\delta
+
O(\|\delta\|^3),
\\
D_{\mathrm{KL}}(x(\delta)\|q)
&=
D_{\mathrm{KL}}(p\|q)
+
g^\top\delta
+
O(\|\delta\|^2).
\end{align*}
Applying the Fisher-constrained solution and using
$g^\top F^\dagger g=\operatorname{Var}_{i\sim p}[r_i]$ yields
\begin{equation*}
U(\varepsilon)
=
\sqrt{
2\varepsilon\,
\operatorname{Var}_{i\sim p}
\left[
\log\frac{p_i}{q_i}
\right]
}
+
O(\varepsilon).
\end{equation*}
Consequently,
\begin{equation*}
\boxed{
s =
\lim_{\varepsilon\downarrow0}
\frac{U(\varepsilon)^2}{2\varepsilon}
=
\operatorname{Var}_{i\sim p}
\left[
\log\frac{p_i}{q_i}
\right]
}.
\end{equation*}

Thus, \(s\) is not merely a quantity produced by the quadratic Fisher approximation. It is the radius-independent coefficient governing the leading-order improvement of the exact constrained problem. At finite \(\epsilon\), however, the exact utility \(U(\epsilon)\) depends on the chosen radius and need not preserve the ranking induced by \(s\).

\section{Additional Experimental Studies}
\label{app:additional_studies}

This appendix collects the controlled studies that complement the main end-to-end comparison. Unless stated otherwise, the student, teacher, data,
training budget, generation settings, and paired bootstrap procedure match
Section~\ref{sec:experiments}.

\subsection{Reverse-KL Evaluation Results Across Model Pairs}
\label{app:main_table_exps}

\paragraph{Evaluation.}
We sample at temperature $0.7$ and top-$p=0.95$. For each problem, we
generate $K_b=16$ responses on AIME24, AIME25, AMC23, and BRUMO25,
and $K_b=8$ responses on MATH-500, Minerva, and OlympiadBench.
The accuracy of problem $i$ in benchmark $b$ is the fraction of its
$K_b$ responses that are correct, denoted $a_{b,i}$. We report
\begin{equation}
\operatorname{Macro}
=
\frac{1}{7}\sum_{b=1}^{7}
\frac{1}{n_b}\sum_{i=1}^{n_b}a_{b,i},
\qquad
\operatorname{Pooled}
=
\frac{1}{1620}\sum_{b=1}^{7}
\sum_{i=1}^{n_b}a_{b,i},
\end{equation}
where $n_b$ is the number of problems in benchmark $b$.
Thus, Macro weights benchmarks equally, while Pooled weights
problems equally regardless of their number of sampled responses.
We estimate aggregate uncertainty and paired contrasts using
20,000 paired bootstrap resamples over problems.

\paragraph{Baselines.}
Uniform OPD uses the same candidate-set construction, renormalized
reverse-KL loss, and training setup as FUW, with $w_t=1$ at every
position. Each candidate set is the union of the student's top-16 and
teacher's top-16 tokens. On average we observe the union consists of about 19 tokens.
We also compare against TIDE using its native support and TIP Soft-OR
at 50\% retention using FUW's candidate-set construction.


End-to-end mathematical reasoning results for the different model pairs are shown in Tables \ref{tab:end_to_end}, \ref{tab:qwen3_4b_30b_results} and \ref{tab:qwen3_1r7b_14b_results}. Macro averages benchmarks equally; Pooled averages all 1,620 problems.
Best trained-student result in each column is \textbf{bold}.

\begin{table*}[ht]
\centering
\caption{Qwen3-1.7B-Base distilled from Qwen3-8B
}
\label{tab:end_to_end}

\scriptsize
\renewcommand{\arraystretch}{1.08}
\setlength{\tabcolsep}{1.8pt}

\begin{tabular*}{\textwidth}{
@{\extracolsep{\fill}}
lccccccccc
@{}
}
\toprule
\textbf{Method}
& \textbf{AIME24}
& \textbf{AIME25}
& \textbf{AMC23}
& \textbf{MATH-500}
& \textbf{Minerva}
& \textbf{Olymp.}
& \textbf{BRUMO25}
& \textbf{Avg@8 Macro}
& \textbf{Avg@8 Pooled} \\
\midrule

\rowcolor{refbg}
Student (Base)
& $2.3{\pm}1.6$
& $0.4{\pm}0.4$
& $7.4{\pm}2.1$
& $17.5{\pm}1.1$
& $4.7{\pm}0.9$
& $7.6{\pm}0.6$
& $2.9{\pm}1.8$
& $6.1{\pm}0.5$
& $9.9{\pm}0.5$ \\

\rowcolor{refbg}
Teacher (8B)
& $30.6{\pm}7.4$
& $20.6{\pm}7.1$
& $62.3{\pm}4.5$
& $83.9{\pm}1.4$
& $28.4{\pm}2.5$
& $47.7{\pm}1.7$
& $25.8{\pm}7.3$
& $42.8{\pm}1.9$
& $55.1{\pm}1.1$ \\

\midrule

OPD
& $8.8{\pm}4.4$
& $5.4{\pm}3.2$
& $32.0{\pm}3.9$
& $61.6{\pm}1.7$
& $16.8{\pm}1.7$
& $25.6{\pm}1.4$
& $9.4{\pm}4.8$
& $22.8{\pm}1.2$
& $34.5{\pm}1.0$ \\


TIDE
& $11.0{\pm}4.5$
& $2.3{\pm}1.5$
& $33.9{\pm}3.9$
& $62.7{\pm}1.7$
& $16.5{\pm}1.8$
& $25.9{\pm}1.4$
& $8.3{\pm}4.4$
& $22.9{\pm}1.1$
& $35.0{\pm}1.0$ \\

TIP (Soft-OR, 50\%)
& $10.6{\pm}4.8$
& $5.0{\pm}3.4$
& $34.9{\pm}4.3$
& $66.9{\pm}1.7$
& $19.5{\pm}2.0$
& $29.2{\pm}1.5$
& $11.0{\pm}5.3$
& $25.3{\pm}1.4$
& $38.4{\pm}1.0$ \\

\midrule

\rowcolor{fuwbg}
\textcolor{fuwgreen}{\textbf{FUW (Ours)}}
& $\mathbf{\textcolor{fuwgreen}{12.7{\pm}4.9}}$
& $\mathbf{\textcolor{fuwgreen}{8.5{\pm}4.1}}$
& $\mathbf{\textcolor{fuwgreen}{37.0{\pm}4.4}}$
& $\textcolor{fuwgreen}{67.5{\pm}1.7}$
& $\mathbf{\textcolor{fuwgreen}{20.3{\pm}2.1}}$
& $\mathbf{\textcolor{fuwgreen}{29.7{\pm}1.5}}$
& $\textcolor{fuwgreen}{11.9{\pm}5.1}$
& $\mathbf{\textcolor{fuwgreen}{26.8{\pm}1.4}}$
& $\mathbf{\textcolor{fuwgreen}{39.2{\pm}1.1}}$ \\

\bottomrule
\end{tabular*}

\end{table*}

\begin{table*}[ht]
\centering
\caption{
Qwen3-4B-Base distilled from Qwen3-30B-A3B-Instruct-2507
}
\label{tab:qwen3_4b_30b_results}
\resizebox{\textwidth}{!}{%
\begin{tabular}{lccccccccc}
\toprule
\textbf{Method}
& \textbf{AIME24}
& \textbf{AIME25}
& \textbf{AMC23}
& \textbf{MATH-500}
& \textbf{Minerva}
& \textbf{Olympiad}
& \textbf{BRUMO25}
& \textbf{Avg@8 Macro}
& \textbf{Avg@8 Pooled} \\
\midrule
Student 4B-Base
& $4.4{\pm}2.6$
& $0.0{\pm}0.0$
& $17.4{\pm}3.2$
& $21.1{\pm}1.0$
& $5.8{\pm}0.9$
& $15.1{\pm}0.9$
& $8.8{\pm}3.6$
& $10.4{\pm}0.8$
& $14.9{\pm}0.6$ \\

Teacher 30B-A3B
& $75.2{\pm}6.7$
& $65.4{\pm}7.4$
& $91.6{\pm}2.7$
& $95.8{\pm}0.8$
& $35.2{\pm}2.7$
& $67.6{\pm}1.7$
& $66.0{\pm}6.6$
& $71.0{\pm}1.8$
& $72.2{\pm}1.0$ \\
\midrule

OPD
& $29.8{\pm}7.5$
& $28.7{\pm}7.6$
& $60.5{\pm}4.7$
& $84.4{\pm}1.3$
& $33.3{\pm}2.5$
& $47.6{\pm}1.7$
& $29.6{\pm}7.4$
& $44.8{\pm}2.0$
& $56.2{\pm}1.1$ \\

TIDE
& $31.5{\pm}8.0$
& $25.8{\pm}7.5$
& $59.0{\pm}4.8$
& $\mathbf{86.6{\pm}1.3}$
& $33.2{\pm}2.5$
& $\mathbf{50.3{\pm}1.7}$
& $\mathbf{38.8{\pm}7.4}$
& $\mathbf{46.4{\pm}2.0}$
& $\mathbf{58.0{\pm}1.1}$ \\

TIP
& $\mathbf{32.3{\pm}7.7}$
& $23.5{\pm}7.2$
& $61.9{\pm}4.6$
& $85.2{\pm}1.3$
& $32.3{\pm}2.6$
& $47.3{\pm}1.7$
& $36.5{\pm}7.4$
& $45.6{\pm}2.0$
& $56.3{\pm}1.1$ \\
\midrule

\rowcolor{green!8}
\textbf{FUW (Ours)}
& $26.7{\pm}7.2$
& $\mathbf{32.3{\pm}7.6}$
& $\mathbf{63.4{\pm}4.5}$
& $85.6{\pm}1.4$
& $\mathbf{33.8{\pm}2.6}$
& $49.2{\pm}1.8$
& $32.1{\pm}7.5$
& $46.15{\pm}1.98$
& $57.54{\pm}1.01$ \\
\bottomrule
\end{tabular}%
}
\end{table*}

\begin{table*}[ht]
\label{tab:1.7_14}
\centering
\caption{Qwen3-1.7B-Base distilled from Qwen3-14B}
\label{tab:qwen3_1r7b_14b_results}
\resizebox{\textwidth}{!}{%
\begin{tabular}{lcccccccccc}
\toprule
\textbf{Method} & \textbf{AIME24} & \textbf{AIME25} & \textbf{AMC23} & \textbf{MATH-500} &
\textbf{Minerva} & \textbf{Olympiad} & \textbf{BRUMO25} &
\textbf{Avg@8 Macro} & \textbf{Avg@8 Pooled} \\
\midrule
Student (Qwen3-1.7B-Base)
& $2.3{\pm}1.6$
& $0.4{\pm}0.4$
& $7.4{\pm}2.1$
& $17.5{\pm}1.1$
& $4.7{\pm}0.9$
& $7.6{\pm}0.6$
& $2.9{\pm}1.8$
& $6.1{\pm}0.5$ 
& $9.9{\pm}0.5$ \\

Teacher (Qwen3-14B) & $31.2{\pm}7.2$ & $28.1{\pm}7.6$ & $63.6{\pm}4.4$ & $86.2{\pm}1.3$ & $32.7{\pm}2.7$ & $51.7{\pm}1.7$ & $35.0{\pm}7.9$ & $46.9{\pm}2.0$ & $58.6{\pm}1.0$ \\
\midrule
OPD  & $10.6{\pm}4.7$ & $\mathbf{7.3{\pm}3.8}$ & $31.6{\pm}3.9$ & $62.2{\pm}1.8$ & $18.7{\pm}1.8$ & $26.4{\pm}1.4$ & $10.2{\pm}4.9$ & $23.9{\pm}1.3$ & $35.5{\pm}0.9$ \\
TIDE & $10.0{\pm}4.8$ & $5.0{\pm}3.4$ & $34.4{\pm}4.1$ & $65.6{\pm}1.7$ & $18.8{\pm}1.9$ & $28.3{\pm}1.4$ & $9.4{\pm}4.4$ & $24.5{\pm}1.3$ & $37.4{\pm}0.9$ \\
TIP  & $10.4{\pm}4.8$ & $6.7{\pm}3.2$ & $34.1{\pm}4.2$ & $66.6{\pm}1.7$ & $20.1{\pm}2.0$ & $29.3{\pm}1.5$ & $\mathbf{15.6{\pm}6.2}$ & $26.1{\pm}1.4$ & $38.5{\pm}0.9$ \\
\rowcolor{green!8}
FUW (Ours) & $\mathbf{11.9{\pm}5.1}$ & $7.1{\pm}3.9$ & $\mathbf{36.3{\pm}4.3}$ & $\mathbf{69.4{\pm}1.7}$ & $\mathbf{22.2{\pm}2.0}$ & $\mathbf{30.2{\pm}1.5}$ & $11.9{\pm}4.7$ & $\mathbf{27.0{\pm}1.3}$ & $\mathbf{40.2{\pm}0.9}$ \\
\bottomrule
\end{tabular}}
\end{table*}

\subsection{Token-Importance Agreement Analysis}
\label{app:importance-agreement}

\paragraph{Token-importance criteria.}
Table ~\ref{tab:importance_views} summarizes the criteria considered in our
agreement analysis and their relationship to the reverse-KL Fisher utility.
KL disagreement and the squared Euclidean gradient norm are analytically
related to the same log-ratio underlying FUW, whereas student entropy,
TIP Soft-OR, and TIDE characterize different properties of the position.

\begin{table*}[ht]
\centering
\small
\setlength{\tabcolsep}{5pt}
\begin{tabularx}{\textwidth}{p{2.8cm} p{4.1cm} X}
\toprule
\textbf{Criterion} & \textbf{Score or quantity} & \textbf{Relationship to Fisher utility} \\
\midrule
FUW (ours)
& $\operatorname{Var}_{p}[r]$
& Maximum first-order reverse-KL reduction, normalized for an equal
  infinitesimal change in the predictive distribution. \\[3pt]

KL disagreement
& $D_{\mathrm{KL}}(p\|q)=\mathbb{E}_{p}[r]$
& Mean of the log-ratio whose centered variance defines Fisher utility.
  Similar KL values need not imply identical Fisher utilities. \\[3pt]

Squared Euclidean gradient norm
& $\|g\|_2^2=\sum_i p_i^2(r_i-\bar r)^2$
& Uses the same centered log-ratio deviations, with an additional
  factor of $p_i$ that relatively favors high-probability student
  outcomes. \\[3pt]

Student entropy
& $H(p)=-\mathbb{E}_{p}[\log p_i]$
& Measures student uncertainty without using the teacher distribution. \\[3pt]

TIP Soft-OR
& $\widehat H+\widehat D-\widehat H\widehat D$
& Combines normalized uncertainty and disagreement scores. \\[3pt]

TIDE
& Excess- and deficit-mismatch scores
& Distinguishes the two directions of student--teacher probability
  mismatch. \\
\bottomrule
\end{tabularx}
\caption{Token-importance criteria used in the agreement analysis.
Here $p$ and $q$ denote the student and teacher next-token distributions
at a fixed prefix, $r_i=\log(p_i/q_i)$, and
$\bar r=\mathbb{E}_{i\sim p}[r_i]$.
The first three rows have the stated analytical relationships; the
remaining criteria characterize different properties of the position.}
\label{tab:importance_views}
\end{table*}

\paragraph{Agreement analysis.}
We compute all criteria on checkpoints obtained during uniformly weighted
reverse-KL OPD of Qwen3-1.7B-Base $\leftarrow$ Qwen3-14B. Since none of
the importance criteria determines this training trajectory, it provides
a controlled setting for analyzing criterion agreement.

The main-text analysis summarizes the two directions using bidirectional
weight retention,
\(B_\rho(X)=\min\{\mathrm{REV}_\rho(X),\mathrm{CONT}_\rho(X)\}\).
Here, we decompose \(B_\rho\) into its directional components to determine
whether high effective-support agreement is symmetric or instead reflects
one-sided containment. Importantly, high bidirectional weight retention
measures agreement in effective high-importance support, not equality of
the underlying weight vectors. Two criteria can attain $B_\rho\approx1$
while assigning different relative weights to individual positions within
that shared support. This distinction is reflected in the substantial differences in 
score concentration across criteria shown in Table~\ref{tab:score_concentration}.

\begin{table}[ht]
\centering
\caption{Fraction of positions needed to account for a given fraction of
score mass.}
\label{tab:score_concentration}
\small
\begin{tabular*}{\columnwidth}{@{\extracolsep{\fill}}lccc@{}}
\toprule
\textbf{Score} & \textbf{50\% mass} & \textbf{90\% mass} & \textbf{95\% mass} \\
\midrule
FUW & $0.077$ & $0.328$ & $0.462$ \\
Euclidean $\|g\|_2^2$ & $0.035$ & $0.121$ & $0.168$ \\
Reverse-KL divergence & $0.038$ & $0.231$ & $0.345$ \\
TIP Soft-OR & $0.055$ & $0.317$ & $0.474$ \\
Uniform & $0.500$ & $0.900$ & $0.950$ \\
\bottomrule
\end{tabular*}
\end{table}

Let $S_X^\rho$ denote the top-$\rho$ fraction of positions according to
criterion $X$, and let $w_{X,i}$ denote the corresponding importance
weight assigned to position $i$. As defined in Section \ref{sec:overlap}
\[
\mathrm{REV}_\rho(X)
=
\frac{
\sum_{i\in S_X^\rho} w_{\mathrm{FUW},i}
}{
\sum_{i\in S_{\mathrm{FUW}}^\rho} w_{\mathrm{FUW},i}
},
\]
measures how much of FUW's attainable top-$\rho$ weight is retained
when positions are instead selected by $X$, and
\[
\mathrm{CONT}_\rho(X)
=
\frac{
\sum_{i\in S_{\mathrm{FUW}}^\rho} w_{X,i}
}{
\sum_{i\in S_X^\rho} w_{X,i}
},
\]
measures the converse. Their difference,
\[
A_\rho(X)
=
\mathrm{CONT}_\rho(X)
-
\mathrm{REV}_\rho(X),
\]
characterizes directional asymmetry. Positive values indicate that FUW's
selected set captures more of $X$'s attainable weight than $X$'s selected
set captures of FUW's; negative values indicate the reverse. Values near
zero indicate approximately symmetric retention.

\paragraph{Random selection controls.}

To distinguish criterion agreement from retention expected under random
selection, we use a permutation control that preserves the weight multiset
while destroying the correspondence between positions and scores. Let
\[
D_\rho =
\frac{\sum_{i\in S^\rho_{\mathrm{FUW}}} w_{\mathrm{FUW},i}}
     {\sum_i w_{\mathrm{FUW},i}}
\]
denote the fraction of FUW's total weight contained in its top-$\rho$
positions. Since each position is included in a uniformly random
top-$\rho$ set with probability $\rho$,
\[
\mathbb{E}[\mathrm{REV}_{\rho,\mathrm{random}}]
=
\frac{\rho}{D_\rho}.
\]
Thus, the control accounts for retention induced by the selection
fraction and the concentration of FUW's weight distribution. The measured
permutation control agrees with this analytic expectation to within
$0.003$ at all 13 checkpoints.

Figures ~\ref{fig:containment-rho02} and \ref{fig:containment-rho05} report the decomposition of $B_{\rho}$ at
$\rho=0.2$ and $\rho=0.5$, corresponding to the 20\% and 50\% retention
fractions. The former is close to TIDE's realized
active fraction in our implementation, while the latter matches the
50\% Soft-OR configuration recommended by TIP.

At $\rho=0.2$, the two directions can differ materially for student entropy and TIP Soft-OR, with asymmetry reaching
approximately $0.18$ over the latter half of training. These differences
coincide with changes in the relative concentration of the criteria's
weights, so we do not interpret them as evidence of directional
containment.

At $\rho=0.5$, Euclidean gradient magnitude, student entropy, KL disagreement, TIP Soft-OR, and TIDE-deficit
exhibit nearly symmetric agreement with FUW: their two directional
retention scores remain close throughout the later checkpoints. Their
near-unit $B_{0.5}$ values therefore reflect high retention in both
directions. TIDE-excess exhibits a qualitatively different pattern. Its asymmetry is
large and positive, indicating that FUW's top-half set captures
substantially more of TIDE-excess's attainable weight than the
TIDE-excess set captures of FUW's. By the later checkpoints, FUW's
selected set captures nearly all of the high-weight support identified by
TIDE-excess, while the converse retention is substantially lower. Thus, unlike the near-symmetric agreement observed for the other criteria,
the lower $B_{0.5}$ of TIDE-excess reflects asymmetric effective-support
containment. This behavior is consistent with the distinct role of
TIDE-excess, which explicitly isolates one direction of student--teacher
mismatch rather than measuring scalar disagreement, uncertainty, or local
optimization opportunity.

\begin{figure}[t]
    \centering
    \includegraphics[width=\textwidth]
    {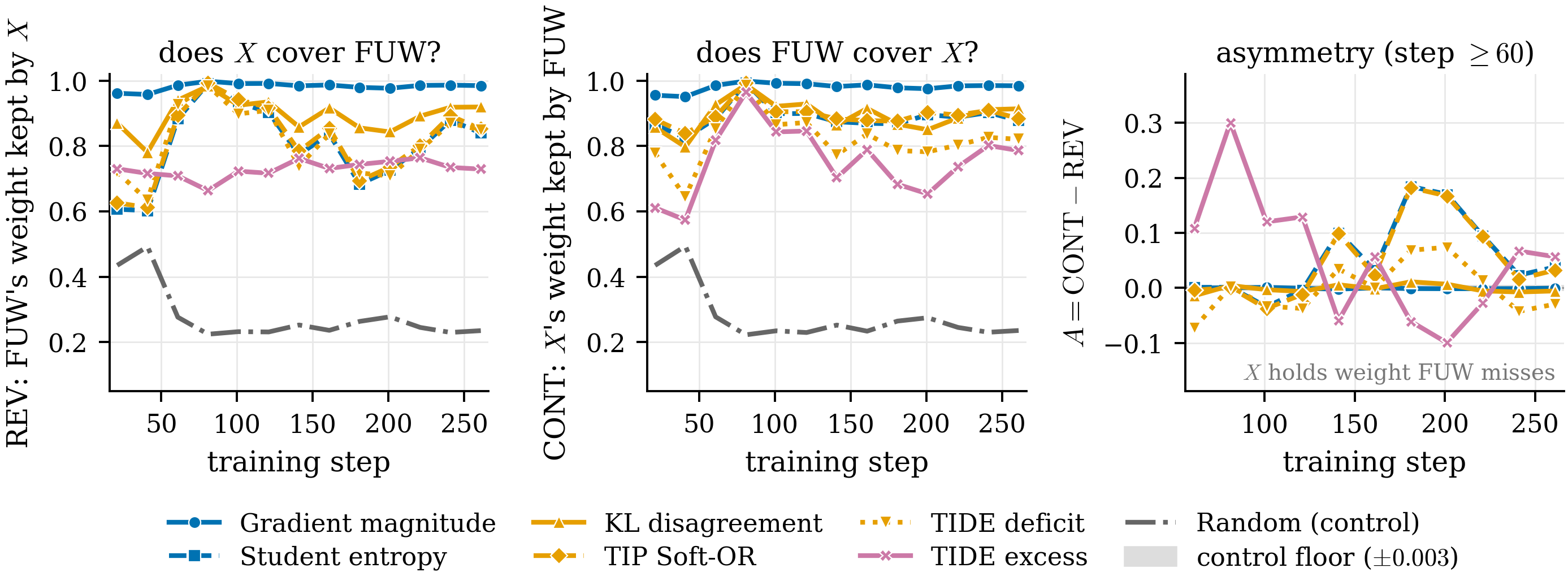}
    \caption{
    \textbf{Directional decomposition at $\rho=0.2$.}
    REV measures the fraction of FUW's attainable top-$\rho$ weight
    retained when positions are selected by criterion $X$, while CONT
    measures the converse. Directional agreement is more sensitive to
    differences in weight concentration at this operating fraction.
    }
    \label{fig:containment-rho02}
\end{figure}

\begin{figure}[t]
    \centering
    \includegraphics[width=\textwidth]
    {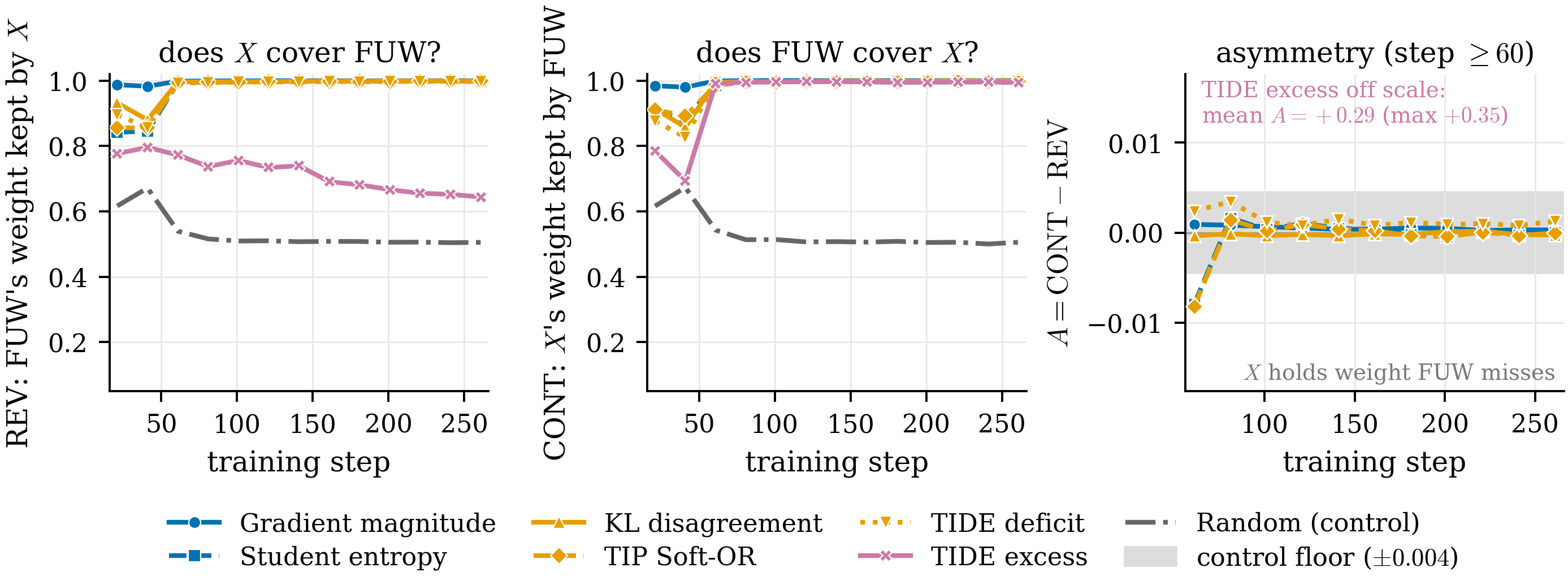}
    \caption{
    \textbf{Directional decomposition at $\rho=0.5$.}
    Most criteria exhibit nearly symmetric, near-complete retention
    with FUW, whereas TIDE-excess exhibits asymmetric effective-support
    containment.
    }
    \label{fig:containment-rho05}
\end{figure}

Overall, for the Qwen3-1.7B-Base $\leftarrow$ Qwen3-14B pair studied
under uniform OPD, several independently motivated criteria concentrate
importance on largely the same effective support as the Fisher-derived
utility at $\rho=0.5$. For gradient magnitude, student entropy, KL
disagreement, TIP Soft-OR, and TIDE-deficit, this relationship is nearly
symmetric, indicating shared effective high-importance support rather
than one-sided containment. TIDE-excess is the notable exception: its
high-importance support is largely contained within FUW's broader
support, while the converse does not hold. These results concern
effective support rather than point-wise equivalence of the weighting
rules.

\paragraph{Cross-pair robustness.}

We additionally evaluate bidirectional weight retention on a second
student--teacher pair using checkpoints generated during FUW training.
Unlike the uniformly weighted trajectory above, these checkpoints are
not independent of the reference criterion, since FUW itself determines
the training trajectory. We therefore treat this experiment as a
cross-pair robustness check rather than an independent test of criterion
agreement.

The similar qualitative structure observed earlier is preserved across both
operating fractions as shown in Figure \ref{fig:wt_equiv_fuw}. At $\rho=0.5$, gradient magnitude, student entropy,
KL disagreement, TIP Soft-OR, and TIDE-deficit exhibit near-complete
bidirectional retention with FUW throughout most of training, whereas
TIDE-excess remains substantially lower. At $\rho=0.2$, the same
separation emerges after the earliest checkpoints: the scalar criteria
approach high effective-support agreement with FUW, while TIDE-excess
remains distinctly separated. In both cases, agreement is substantially
above the random-selection control. Thus, the effective-support structure
observed under uniform OPD for the primary model pair is qualitatively
reproduced on a different model pair and training trajectory.

\begin{figure*}[t]
    \centering
    \includegraphics[width=\textwidth]{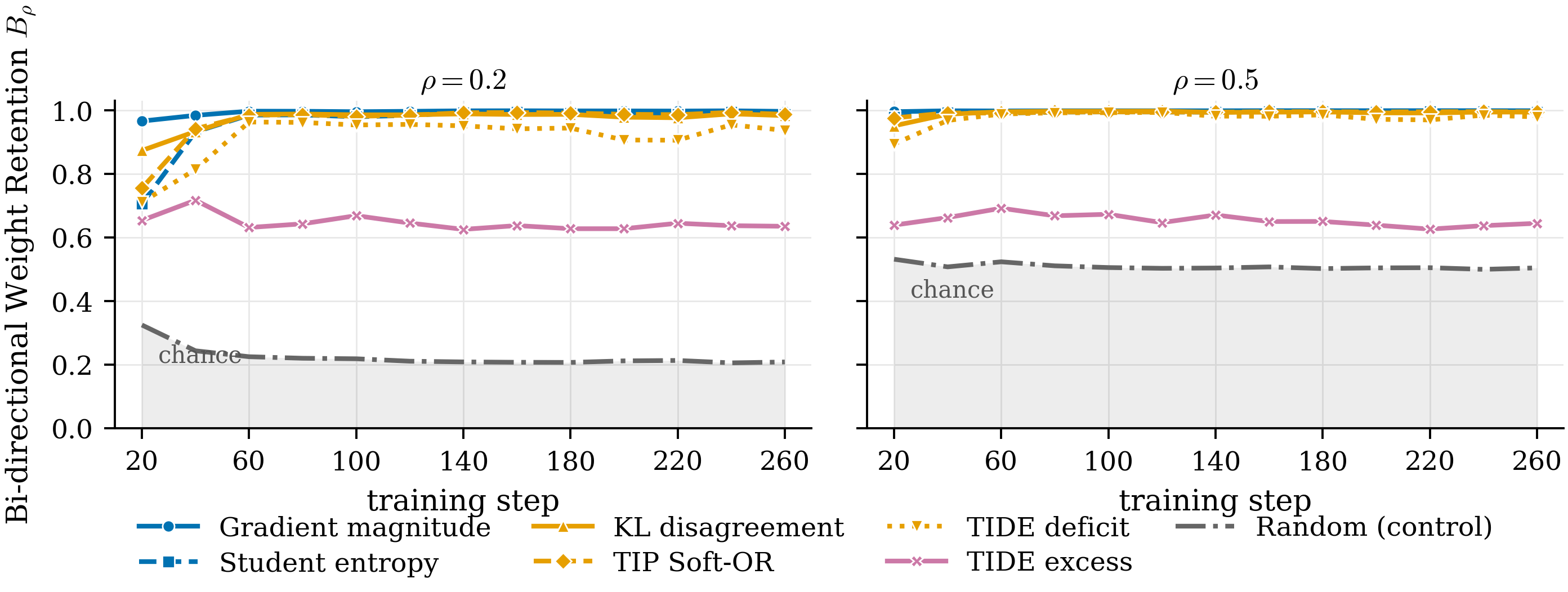}
    \caption{Cross-pair robustness of effective-support agreement. Bidirectional weight retention \(B_\rho\) with FUW on Qwen3-1.7B-Base \(\leftarrow\) Qwen3-8B along an FUW-trained trajectory, at \(\rho=0.2\) (left) and \(\rho=0.5\) (right). Beyond the early training steps, most criteria show near-complete effective-support agreement with FUW at both retention fractions, while TIDE-excess remains distinctly lower.}
    \label{fig:wt_equiv_fuw}
\end{figure*}

\subsection{Weighting, metric, and ranking controls}
\label{app:weighting_controls}

All controls in this subsection use the same re-normalized union of the student's and teacher's top-16 candidates obtained during the reverse-KL distillation of the Qwen3-1.7B-Base \(\leftarrow\) Qwen3-8B pair. They therefore isolate the scoring or weighting rule without changing 
candidate support. Table~\ref{tab:weighting_controls} reports paired pooled-score differences relative to FUW. A negative value favors FUW.

\begin{table}[ht]
\centering
\caption{Paired controls on the union top-16 support for reverse-KL OPD of Qwen3-1.7B-Base $\leftarrow$ Qwen3-8B pair. The Euclidean variant
replaces the Fisher utility by $\|g_t\|_2^2$; the permutation control randomly
reassigns FUW weights within each batch. Negative $\Delta$ favors FUW.}
\label{tab:weighting_controls}
\setlength{\tabcolsep}{4pt}
\begin{tabular*}{0.82\linewidth}{@{\extracolsep{\fill}}lcccc@{}}
\toprule
\textbf{Method}
& \textbf{Pooled Avg@8}
& $\boldsymbol{\Delta}$ \textbf{vs.\ FUW}
& \textbf{95\% CI}
& $\boldsymbol{p}$ \\
\midrule
FUW
& 39.17
& --
& --
& -- \\

Euclidean utility
& 38.9
& $-0.27$
& $[-1.09,\ 0.56]$
& $0.527$ \\

Permuted FUW
& 38.15
& $-1.02$
& $[-1.92,\ -0.11]$
& \textbf{0.028} \\

Hard top-half FUW
& 38.41
& $-0.76$
& $[-1.62,\ 0.11]$
& $0.087$ \\
\bottomrule
\end{tabular*}
\end{table}

The Euclidean variant reaches pooled Avg@8 of \(38.9\), close to FUW's \(39.2\), with no significant paired difference (\(p=0.527\)). However, its weighting dynamics differ substantially from FUW. Effective retention decreases from \(0.327\) early in training to \(0.104\) by step 118, while the maximum token weight rises from \(29\) to \(96\). The additional factor of \(p_S\) implicit in the Euclidean score increasingly concentrates weight on high-probability student outcomes. Fisher normalization avoids this late-stage concentration, although the two metrics are not separated by downstream accuracy at the present training budget.

The permutation control preserves the complete multiset of FUW weights and
changes only their assignment to positions. Its significant reduction
($-1.02$, $p = 0.028$) shows that position-specific utility information
contributes beyond the marginal weight distribution: the positions FUW
identifies are not interchangeable. Hard selection does the opposite,
preserving which positions FUW identifies as high utility while discarding
their relative weights. Used as a top-half mask, the FUW score reaches
$38.4$ pooled Avg@8, matching TIP Soft-OR at the same 50\% retention
fraction ($38.4$). This matches the effective-support analysis in
Section~\ref{sec:overlap}: the two criteria identify nearly the same support
at $\rho = 0.5$. Continuous FUW reaches $39.2$; on this pair, its difference from the hard
mask is not significant ($p = 0.087$).

\subsection{Forward-KL Evaluation Results Across Model Pairs}
\label{app:fkl_results}

Table ~\ref{tab:fkl-pass8} reports the per-benchmark results for the
forward-KL experiments summarized in Section \ref{sec:fkl}.
\begin{table*}[ht]
\centering
\small
\setlength{\tabcolsep}{3.5pt}
\renewcommand{\arraystretch}{1.05}

\resizebox{\textwidth}{!}{%
\begin{tabular}{@{}lccccccccc@{}}
\toprule
\textbf{Method}
& \textbf{AIME24}
& \textbf{AIME25}
& \textbf{AMC23}
& \textbf{MATH}
& \textbf{Minerva}
& \textbf{Olymp.}
& \textbf{BRUMO}
& \textbf{\shortstack{Avg@k\\Macro}}
& \textbf{\shortstack{Avg@k\\Pooled}} \\
\midrule

\multicolumn{10}{@{}l}{
\textit{\textbf{Qwen3-1.7B-Base
$\leftarrow$
Qwen3-8B}}
} \\[2pt]

FKL (Uniform)
& $4.6{\pm}3.2$
& $2.1{\pm}1.5$
& $25.7{\pm}3.7$
& $56.9{\pm}1.7$
& $19.4{\pm}1.9$
& $22.4{\pm}1.3$
& $\mathbf{9.4{\pm}3.1}$
& $20.07{\pm}0.94$
& $31.78{\pm}0.83$ \\

FKL + Utility (Ours)
& $\mathbf{12.1{\pm}4.5}$
& $\mathbf{4.4{\pm}3.0}$
& $\mathbf{28.3{\pm}4.0}$
& $\mathbf{63.6{\pm}1.8}$
& $\mathbf{20.4{\pm}2.1}$
& $\mathbf{24.7{\pm}1.3}$
& $6.0{\pm}3.0$
& $\mathbf{22.78{\pm}1.14}$
& $\mathbf{35.21{\pm}0.89}$ \\

\midrule

\multicolumn{10}{@{}l}{
\textit{\textbf{Qwen3-1.7B-Base
$\leftarrow$
Qwen3-14B}}
} \\[2pt]

FKL (Uniform)
& $3.8{\pm}1.6$
& $0.5{\pm}0.3$
& $\mathbf{29.4{\pm}3.8}$
& $57.7{\pm}1.7$
& $\mathbf{20.0{\pm}1.9}$
& $22.4{\pm}1.3$
& $\mathbf{9.6{\pm}3.9}$
& $20.46{\pm}0.91$
& $32.23{\pm}0.84$ \\

FKL + Utility (Ours)
& $\mathbf{6.9{\pm}3.7}$
& $\mathbf{2.7{\pm}1.7}$
& $27.8{\pm}3.9$
& $\mathbf{63.0{\pm}1.7}$
& $19.1{\pm}2.0$
& $\mathbf{24.0{\pm}1.3}$
& $7.3{\pm}3.5$
& $\mathbf{21.54{\pm}1.02}$
& $\mathbf{34.41{\pm}0.86}$ \\

\midrule

\multicolumn{10}{@{}l}{
\textit{\textbf{Qwen3-4B-Base
$\leftarrow$
Qwen3-30B-A3B-Instruct-2507}}
} \\[2pt]

FKL (Uniform)
& $\mathbf{24.0 {\pm} 5.3}$
& $21.0 {\pm}5.7$
& $\mathbf{60.1 {\pm} 5.2}$
& $84.0 {\pm}1.5$
& $32.0 {\pm}2.7 $
& $48.5 {\pm}1.8$
& $24.8 {\pm}7.4$
& $42.04 {\pm}1.7 $
& $55.85 {\pm}1.2$ \\


FKL + Utility (Ours)
& $22.9{\pm}5.5$
& $\mathbf{22.9{\pm}5.9}$
& $57.0{\pm}5.1$
& $\mathbf{85.5{\pm}1.6}$
& $\mathbf{32.7{\pm}2.9}$
& $\mathbf{47.6{\pm}1.9}$
& $\mathbf{33.5{\pm}7.2}$
& $\mathbf{43.2{\pm}1.8}$
& $\mathbf{56.1{\pm}1.2}$ \\

\bottomrule
\end{tabular}%
}
\caption{
Forward-KL mathematical-reasoning performance across three
student--teacher pairs. Benchmark columns report Avg@k, while aggregate columns report
macro- and pooled-average Avg@k. Values are reported as mean $\pm$ standard error.
The best result within each model pair is shown in bold.
}
\label{tab:fkl-pass8}
\end{table*}

\end{document}